\documentclass[10pt,journal]{IEEEconf}
\usepackage{amsmath,amsfonts}

\usepackage{array}
\usepackage[caption=false,font=footnotesize,labelfont=sf,textfont=sf]{subfig}
\usepackage{textcomp}
\usepackage{stfloats}
\usepackage{url}
\usepackage{verbatim}
\usepackage{graphicx}
\usepackage{amssymb}
\usepackage{pifont}
\usepackage{siunitx}
\usepackage{algorithm,algcompatible}
\usepackage{todonotes}
\usepackage{hyperref}
\usepackage[capitalize]{cleveref}
\usepackage{comment}
\usepackage{multirow}
\usepackage{tikz}
\usepackage{pgfplots}

\algnewcommand\FUNC{\item[\textbf{Function Signature:}]}
\algnewcommand\INPUT{\item[\textbf{Input:}]}
\algnewcommand\OUTPUT{\item[\textbf{Output:}]}
\algnewcommand\RETURN{\STATE{\textbf{return\ }}}

\begin{document}

\title{\LARGE \bf A new Taxonomy for Automated Driving: Structuring Applications based on their Operational Design Domain, Level of Automation and Automation Readiness}
\author{Johannes Betz$^{1}$, Melina Lutwitzi$^{2}$, Steven Peters$^{2}$  
\thanks{$^{1}$J. Betz is with the Professorship of Autonomous Vehicle Systems, TUM School of Engineering and Design, Technical University Munich, 85748 Garching, Germany; Munich Institute of Robotics and Machine Intelligence (MIRMI), \{{johannes.betz}\}@tum.de}
\thanks{$^{2}$S. Peters as full professor and M. Lutwitzi as chief engineer are with the Institute of Automotive Engineering at TU Darmstadt; Department of Mechanical Engineering, 64289 Darmstadt, Germany; \{{steven.peters}\}@tu-darmstadt.de}
}



\maketitle
\thispagestyle{empty} 
\pagestyle{empty}    

\begin{abstract}

The aim of this paper is to investigate the relationship between operational design domains (ODD), automated driving SAE Levels, and Technology Readiness Level (TRL). The first highly automated vehicles, like robotaxis, are in commercial use, and the first vehicles with highway pilot systems have been delivered to private customers. 
It has emerged as a crucial issue that these automated driving systems differ significantly in their ODD and in their technical maturity. Consequently, any approach to compare these systems is difficult and requires a deep dive into defined ODDs, specifications, and technologies used. Therefore, this paper challenges current state-of-the-art taxonomies and develops a new and integrated taxonomy that can structure automated vehicle systems more efficiently. We use the well-known SAE Levels 0-5 as the "level of responsibility", and link and describe the ODD at an intermediate level of abstraction.  Finally, a new maturity model is explicitly proposed to improve the comparability of automated vehicles and driving functions. This method is then used to analyze today's existing automated vehicle applications, which are structured into the new taxonomy and rated by the new maturity levels. Our results indicate that this new taxonomy and maturity level model will help to differentiate automated vehicle systems in discussions more clearly and to discover white fields more systematically and upfront, e.g. for research but also for regulatory purposes.

\end{abstract}



\section{Introduction}

Automated driving systems (ADS) have made significant strides across various domains. Most of these systems are categorized with the help of the taxonomy of SAE Levels \cite{SAE_Level} (see center in Figure 1), which categorizes automation in vehicles from Level 0 (no automation) to Level 5 (full automation). Several local tests with peoplemover and automated shuttles have been showcased, while all major OEMs offer SAE Level 2 systems in series production cars. The industry is actively exploring Level 3 automation, which permits conditional automation, enabling the driver to disengage temporarily in specific scenarios. However, drawbacks include the challenge of regaining control in emergencies and ambiguities in the handover process between the vehicle and the driver \cite{McCall2019,Wintersberger2017,Gluck2022}. Mercedes-Benz was granted approval by German KBA for its SAE Level 3 system called \textit{DrivePilot} based on UN R157 in December 2021 \cite{KBA} and has a certification for selected US markets (e.g., based on Nevada Chapter 482A) since January 2023. At SAE Level 4, high automation enables the vehicle to operate autonomously without a handover claim to a driver, but not on all roads or under all conditions. Existing systems are operated in predefined areas, such as closed campuses or dedicated lanes. This technology shows promise in controlled environments but struggles to handle unexpected situations or off-road scenarios. Companies like Waymo have been offering their Level 4 robotaxis services in San Francisco as a commercial service since August 2023, after years of piloting in Phoenix and San Francisco.

It becomes clear that up to SAE Level 4, an associated definition of the scope of the operational design domain ODD (see top in Figure 1) is essential to describe or differentiate systems. Only at Level 5 an "unlimited" ODD can be assumed. In addition, different levels of maturity become apparent. One system already has an operating license, while another is only authorized by a safety driver or on representative testing grounds. The technical maturity depends on the degree of verification and validation in the target environment and can be expressed by the technology readiness level (TRL) (see bottom in Figure 1). 

\begin{figure}
    \centering
    \includegraphics{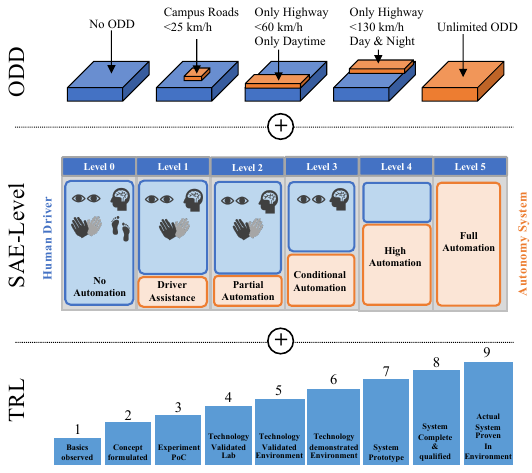}
    \caption{An illustrative comparison of distinct ODDs, the prevalent SAE Levels \cite{SAE_Level} in automated driving, and TRLs.}
    \label{fig:enter-label}
\end{figure}

 The SAE taxonomy was introduced to define the automation level of a vehicle in terms of the driver's remaining driving tasks and responsibilities. As a very general framework, it does not cover any information about the limitations of the system's operation. A vehicle rated at a certain level may operate well in one location but struggle in a different environment with unique challenges. For instance, one vehicle with SAE Level 4 might be capable of driving automated at night, and another, rated as Level 4, might not. Moreover, a one-of-a-kind research vehicle with Level 4 might be capable of driving automation in a whole city based on a safety concept that is not scalable to broader cases.
 
 It becomes apparent that the currently widely used SAE Level categorization is insufficient to structure ADS and categorize them according to their actual technical capabilities and maturity. We, therefore, want to provide a new type of taxonomy that helps to categorize an ADS's capabilities and maturity with greater clarity. The final result shall support science and industry, regulators, and mass media in comparing approaches and identifying white spots. 
 
 In summary, we combine new ODD definitions with the technical capabilities of an automated vehicle (AV). Therefore, this work contributes to a new taxonomy for ADS that extends the well-known responsibility levels with a new intermediate-level ODD structure and a TRL for autonomous systems. For the latter, we suggest our newly created “Autonomous Readiness Levels”.
 


\section{Related Work}\label{chapter:relatedwork}

\textbf{The SAE Levels:}
In 2021, SAE and ISO updated SAE J3016 \cite{SAE_Level}, which describes the levels of driving automation. Where Level 0 describes features for warning or momentary assistance (such as automatic emergency braking), Level 1 systems provide steering or breaking/acceleration support, whereas Level 2 systems provide both (e.g., lane centering and adaptive cruise control). Starting with Level 3, the human driver is not driving anymore when systems are engaged. However, Level 3 needs a human driver who must drive whenever the system requests, whereas Levels 4 and 5 will not request to take over the driving task. The only difference between Level 4 and Level 5 is that Level 4 might be restricted to certain limited conditions (e.g. local robotaxi) and might not operate unless all requirements are met. Although this taxonomy is heavily used, especially due to legal reasons, researchers \cite{Hopkins2021Talking,Richardson2021} made a critical reflection on the use of SAE levels of automation.

\textbf{Taxonomies for ODDs:}
In industry and science, the term \textit{operational design domain} is of tremendous importance when defining or comparing ADS. ISO/FDIS 34503:2023 \cite{ISO_ODD} defines ODD as “the operating conditions within which an ADS can perform the dynamic driving task safely during a trip.” It, therefore, defines the boundaries within which the ADS will function correctly with the given technical capabilities. Consequently, an ODD must cover different ontologies like road types, other participants, environmental conditions (e.g., weather), sight conditions, etc \cite{Chen2022, Won2020, Czarnecki2018}. ISO with 34503, BIS with PAS 1883:2020 \cite{BSI_PAS1883}, and ASAM with OpenODD define taxonomies and formats to describe an ODD in a standardized way. All of these approaches are highly important, especially to specify an ADS or a safety approval for ADS as complete and detailed as possible \cite{Weissensteiner2023,Ito2021}. However, these frameworks are not made for and not suitable for structuring existing ADS in science and industry, identifying white spots in regulation, or finding potential opportunities for technology or business development. 

\textbf{Technology Readiness Level (TRL):}
In general, the maturity or readiness of any technology is frequently rated using derivatives based on the Technology Readiness Level model by NASA. The original NASA model as published in \cite{Mankins1995} starts with Level 1 where basic principles are reported and ends up with Level 9 where an actual system was proven in flight. To the authors' knowledge, no readiness-level model dedicated to automated driving technologies nor systems exists.

\textbf{New Taxonomy Approaches for AVs}
The need for new taxonomies arises from the limitations of existing frameworks in capturing the complexity of modern autonomous systems. The CEO and CTO of Mobileye \cite{mobileyeDefiningTaxonomy} propose a new taxonomy for consumer AV as a substitute of the current SAE Levels. They argue that the SAE levels, originally meant to clearify the meaning of "autonomous driving" for developers and regulations, is not suitable for a "product-oriented description that is clear for both the engineer and the end customer", as it cannot define how the ADS "will work in the real world and ensures the usefulness, safety, and scalability". Besides unclear definitions for the end-user, they criticize the "unnecessary distinction between Level 3 and Level 4, which differs in the Minimum Risk Maneuver (MRM) requirements and the vigilance level of the human driver", but can, in their opinion, lead to failures by design. To address the deficiencies, they propose "a simplified language that defines the levels of autonomy based on four axes: (i) Eyes-on/Eyes-off, (ii) Hands-on/Hands-off, (iii) Driver versus No-driver, and (iv) MRM (minimal risk maneuver) requirement". The ODD and the technological readiness, however, are not addressed by this new taxonomy. 

Chen et al. \cite{Chen2023} define a new taxonomy for AV considering ambient road infrastructure. The work also criticizes the lack of ODD characterization within the SAE taxonomy to clearly define expectations towards the capabilities and performance of an AV. Therefore, a supplement to the existing SAE levels of automation from a road infrastructure perspective is proposed via five sub-levels for Level 4 automated driving systems: Level 4-A (Dedicated Guideway Level), Level 4-B (Expressway Level), Level 4-C (Well-Structured Road Level), Level 4-D (Limited-Structured road Level), and Level 4-E (Disorganized Area Level).  
Furthermore, Kugele et al. \cite{Kugele2021Towards} present a four-level taxonomy for autonomous systems, comparing it to the SAE J3016 standard, and Gamer et al. \cite{Gamer2020The} proposed a six-level autonomy framework for the process industry, considering cognitive capabilities and AI technologies.

Warg et al. \cite{Warg2023} criticize that current taxonomies do not integrate different types of AV besides regular road vehicles and further only focus on vehicles with individual strategies, meaning without cross-vehicle or infrastructure interaction. They therefore "review and extend taxonomies and definitions to encompass individually acting as well as cooperative and collaborative AVs for both on-road and off-road use cases. In particular, [they] introduce classes of collaborative vehicles not defined in existing literature, and define levels of automation suitable for vehicles where automation applies to additional functions in addition to the driving task." 

Furthermore, various taxonomies are proposed to categorize the levels of automation in different domains, recognizing the shifts in human-system interactions due to automation. For instance, Machado et al. \cite{Machado2021Towards} propose a taxonomy for heavy-duty mobile machinery integrating manipulation and driving operations, extending the widely recognized SAE J3016 levels in the automotive industry. Williams \cite{Williams2021Automated} emphasize the adoption of the J3016 taxonomy in automated driving systems, highlighting its discrete and exclusive levels. Understanding the Operational Design Domains (ODD) is crucial for developing ADS. Ishigooka et al. \cite{Ishigooka2019Graceful} focus on graceful degradation design in ADS for enhancing safety and availability, particularly at Level 3 automation. Torkjazi and Raz \cite{Torkjazi2022A} propose a taxonomy for systems of autonomous systems based on the level of autonomy, using the OODA loop as a fundamental abstraction.

This number of various new taxonomies highlights the need for a common and new categorisation method for ADS. However, existing approaches do not yet provide taxonomies that comprehensively map the ODD at an intermediate level of abstraction, rather than looking at specific aspects such as road infrastructure or collaboration, while technical maturity is not addressed at all.



\begin{table*}[t]
  \centering
  \begin{tabular}{p{2cm} |p{2cm}p{2cm}p{2cm}p{2cm}p{2cm}p{2cm}}
    \textbf{Categories} &  &  &  &  \textbf{Attributes} \\

    \hline
    Country Code & $\star$: any country & \textbf{DE:} Germany & \textbf{JP:} Japan & \textbf{US:} USA  & etc.& \\
    
    \hline
    Road Users & $\star$: anything in mixed traffic &	\textbf{A:} only automated traffic	& \textbf{P:} mixed traffic without VRU & & & \\

    \hline
     Road Types & $\star$: any type of road &	\textbf{H:} highway	(without construction sites) & \textbf{H+:} highway incl. driveway, departure, highway service area, construction site & \textbf{U:} urban & \textbf{C:} country incl. roads without lane markings & \textbf{S:} special road e.g. with special markers \\


    \hline
    Environmental Conditions & $\star$: all weather and light conditions / not applicable & \textbf{L}: only daylight, \hspace{0.3cm} \hspace{1cm} \textbf{N}: also night & \textbf{D:} only dry conditions, \hspace{0.3cm} \hspace{1cm} \textbf{R:} also wet con- ditions (no frost), \hspace{0.3cm} \hspace{1cm} \textbf{I:} also ice/snow  & \textbf{F:} fog  \\

    \hline
    Velocity & $\star$: no limit &	\textbf{v0:} $<$ 7 km/h & \textbf{v1:} $<$ 12 km/h & \textbf{v2:} $<$ 25 km/h & \textbf{v3:} $<$ 60 km/h & \textbf{v4:} $<$ 130 km/h \\

    \hline
    Additional Requirements & any additional requirements e.g. vehicle ahead  & & & \\
    \hline   \\
  \end{tabular}
  \caption{
  Structuring Operational Design Domains with a
Intermediate Level Taxonomy by defining Categories and their attributes for the ADS.}
  \label{tab:1}
\end{table*}

\section{Methodology}\label{chap:meth}
Based on the state of the art and a critical reflection on the use of SAE levels of automation, we want to emphasize a  techno-centric view on today's and future autonomous driving systems. This allows us to shift the focus from a driver-centric model (like the SAE levels) to a technology-oriented framework. We therefore present a three-fold method that consists of three steps to create a new taxonomy for AVs. This will ensure that the new taxonomy is scalable and adaptable to future technological innovations, for example allowing new business models for AVs to categorise their technology.
In the following, we set out the procedure for deriving the new taxonomy proposal and describe the resulting method.

\subsection{Step 1: Structuring Operational Design Domains with a Intermediate Level Taxonomy}
With the drawbacks of the rich in detail ODD definition by ISO 34503, we want to provide a simple taxonomy for ODD on an intermediate level of abstraction, as the final result shall support science and industry as well as regulators and mass media to compare approaches and identify white spots. The ODD taxonomy shall therefore be able to particularly emphasise ODD differences arising from the differing use cases or technical capabilities of systems while reducing the amount of categories and attributes to a neccessary minimum. In addition, it should use terms that are understandable to the general public. In order to identify a suitable ODD description method that has both suitable categories and the necessary granularity of attributes for the intended purpose, the following steps are carried out:

 In the first step, the categorisations from the ODD description according to ISO 34503 are consulted. On the first level, these include \textit{scenery elements}, \textit{environmental conditions} and \textit{dynamic elements} with further subcategories. The next two levels of subcategories are filtered and clustered to such categories that (1) have a strong use case specific dependency (e.g. driving on specific road types) and (2) have a particularly large influence on technical capabilities (e.g. environment perception under certain environmental conditions). In the next step, the results are adapted to known ODD limitations of systems in development or on the market and their use cases to ensure that their key attributes can be mapped in the categories and differences can become visible. Thereby, those categories that only appear in single cases and are not expected to gain in relevance in the future are eliminated. This procedure results in five categories with pre-defined attributes to choose from and one “open” category for the individual aspects. The final result is summarized in \cref{tab:1}. The attributes are explained in the following:


\begin{enumerate}
    \item \textbf{Country code} is derived from \textit{scenery elements - zone} and describes the country specific market for which the system is developed or in which country it is allowed to operate. This category is required because systems currently on the market are only authorised in certain countries. The attributes are described by the standardized country codes according to ISO 3166.
    \item \textbf{Road users} is derived from \textit{dynamic elements - traffic agents} and defines traffic participants with potential influence on the ADS. 
        \begin{itemize} \item \textbf{$\star$} includes all types of potential road users in mixed traffic, meaning automated and non-automated vehicles, motorcycles, bicycles and pedestrians as vulnerable road users (VRU). 
        \item \textbf{$A$} includes only automated traffic and no VRU
        \item \textbf{$P$} includes mixed traffic and no VRU
        \end{itemize}
    \item \textbf{Road types} is derived from \textit{scenery elements - drivable area} and directly attributes common drivable road types of major countries. Other drivable area attributes such as lane markings, edges, signs and surfaces found in ISO 34503 are categorised indirectly by assuming that if a road type is drivable, all specifications of the other attributes that usually occur for this type are covered as well.
        \begin{itemize} 
        \item \textbf{$\star$} includes all road types, lane markings, surfaces, edges, geometries and road signs (for the specific country).  
        \item \textbf{$H$} includes only highways and corresponding further road attributes without construction sites.
        \item \textbf{$H+$} extends the road type "highway" by driveway and departure, construction sites as well as driveway service areas.
        \item \textbf{$U$} covers urban road types with or without lane markings and corresponding further road attributes.
        \item \textbf{$C$} covers country road types and corresponding further road attributes.
        \item \textbf{$S$} covers special road types and corresponding special lane markings (e.g. parking garages), surfaces, edges, geometries and road signs.
        \end{itemize}
    \item \textbf{Environmental conditions} are derived from \textit{environmental conditions - weather} and \textit{illumination} whereas the focus is on conditions effecting sight (e.g., daylight, fog, snow) and vehicle dynamics (e.g., ice), while wind and temperature are neglected.
        \begin{itemize} \item \textbf{$\star$} includes all potential weather and light conditions. 
        \item \textbf{$L$} indicates that daylight is required, respectively night is excluded. 
        \item \textbf{$N$} substitutes L and indicates that night operation is additionally included.
        \item \textbf{$D$} indicates the limitation that dry conditions are required, respectively wet roads and rainfall are excluded.
        \item \textbf{$R$} substitutes D and indicates that wet conditions, meaning wet roads or rainfall are also included, while frost is still excluded.
        \item \textbf{$I$} substitutes R and indicates that wet conditions are extended by ice and snow.
        \item \textbf{$F$} indicates that fog is included in the ODD.
        \end{itemize}
    \item \textbf{Velocity} indicates the velocity limits of the automated vehicle.
        \begin{itemize}
            \item \textbf{$\star$} indicates that no velocity limit exists beyond the prevailing local speed limits. 
            \item \textbf{$v0/v1/v2/v3/v4$} specifies five ascending velocity limits. v0 (7 km/h) is oriented towards walking speed and parking procedures. v1 (12 km/h) is chosen due to relevance for valet parking systems, v2 (25 km/h) from the definition of a generalized low speed automation function according to \cite{Berghoefer23}, v3 and v4 are defined according to the limitations introduced by UN regulation no. 157 for automated lane keeping systems (ALKS).
        \end{itemize}
    \item \textbf{Additional Requirements} provides space for individual restrictions that cannot be accommodated in the above structure.    
\end{enumerate}

To classify a system according to the our ODD taxonomy described above, the different attributes are written in the order of the categories in the following format: \textbf{US $\vert$ $\star$ $\vert$ H+ $\vert$ NR $\vert$ v3 $\vert$ none}

In this example, the attributes can then be connected to the categories and therefore provide a simple taxonomy for the ODD on an intermediate level of abstraction:
\begin{itemize}
    \item Country code: US (USA)
    \item Road user: $\star$ (anything in mixed traffic)
    \item Road types: H+ (highway +)
    \item Environmental condition: NR (day and night operation, in dry and wet conditions, but no ice, no snow, no fog)
    \item Velocity: v3 ($<$60  km/h)
    \item Additionaly Requirements: None
\end{itemize}

\subsection{Step 2: Adding Levels of Automation to the Taxonomy}

The next step is to categorise the degree of automation and the driver's responsibility when the automated driving function is activated. Chapter II presented the levels according to SAE J3016. They mainly describe “what does the human in the driver’s seat have to do” and in doing so they address the “level of responsibility”. Although we criticise the clarity and meaningfulness of the SAE levels for end users in this paper, we propose them as a metric for the degree of automation within our taxonomy. Our criticism, as well as that of some other authors, mainly relates to the use of SAE levels as the only taxonomy for differentiating AD functions, as it does not cover all areas relevant for a (technical) comparison. For the aspect of responsibility, however, it is a useful and, in particular, established method in the current state of the art, which is also essential for the specification of the legal framework. We therefore consider it to be suitable in combination with our ODD and readiness taxonomy. Nevertheless, we note that there is potential for improvement in the comprehensibility of the description of responsibilities, whereas the method proposed by Mobileye, for example, has advantages for end users.

In step 2 of our method, we combine the SAE Levels with the new classifying system according to our new ODD taxonomy from step 1. We add the SAE Level in front of the new attributes written in the order of the categories in the following format: \textbf{4} $\vert$ US $\vert$ $\star$ $\vert$ H+ $\vert$ IF $\vert$ v3 $\vert$ none 

As level 5 is not expectable in the next decades since it means unlimited driving everywhere, everytime and under every condition, we focus on the currently by far most important levels 2, 3 and 4.

\subsection{Step 3: Estimating AD-Readiness}
Now, we have a well-defined high-level ODD description which together with the standardized SAE level allows for a further analysis of current readiness of technology for a certain purpose or ODD. In close alignment with the well-known TRL model by NASA we defined the new \textit{Automated Driving-Readiness Level (ADRL)} model as follows in \cref{tab:2}. For the sake of completeness, these ADRL are defined equivalently for all TRLs, even though relevance for this taxonomy is only considered from level TRL~3 onwards. ADRL~3 and ADRL~4 describe systems that are already usable in in-the-loop simulations and therefore can be applied and tested in a defined ODD (within a simulated but representative environment). Starting at ADRL~5, a full vehicle is required and with ADRL~6 vehicles on the roads are required. With the strong research efforts in the field of simulations there might be a chance to further increase the span of ADRLs where simulation is sufficient. 

In step 3 of our method, we add the respective ADRL to the end of our attribute sequence:
4 $\vert$ US $\vert$ $\star$ $\vert$ H+ $\vert$ IF $\vert$ v3 $\vert$ none $\vert$ \textbf{ADRL9}.

\begin{table}[t]
\begin{tabular}{p{0.2cm}p{3cm}p{4.0cm}}
\hline
 & \textbf{TRL by NASA} & \textbf{ADRL (ours)} \\
\hline
1 & Basic principles observed and reported & Basic Principles observed and reported (e.g. scientific result on new neural network architecture in fundamental research) \\
\hline
2 & Technology concept and/or application formulated  & Technology concept and/or application formulated \\
\hline
3 & Analytical and experimental critical function and/or characteristic proof-of-concept & Proof-of-concept by SiL testing (software-in-the-loop)\\
\hline
4 & Component and/or breadboard validation in lab environment & Verification by HiL testing (hardware-in-the-loop) \\
\hline
5 & Component and,or breadboard validation in relevant environment	 & Verification by ViL testing (vehicle-in-the-loop) \\
\hline
6 & System/subsystem model or prototype demonstration in a relevant environment (ground or space) & Demonstration in real vehicle with safety driver \\
\hline
7 & System prototype demonstration in space environment & Validation in real vehicle in ODD with safety driver\\
\hline
8 &  Actual system completed and “flight qualified” through test and demonstration (ground or space)	& Approval, certification, homologation for series production or customer operation \\
\hline
9 & Actual system “flight proven” through successful mission operations	
  & System in commercial use without safety driver\\
\hline \\
\end{tabular}
\caption{Derivation of the ADRL from the TRL.}
\label{tab:2}
\end{table}


\section{Taxonomy Application}\label{chapter:results}

The theoretical framework we propose is based on a comprehensive review of existing literature and current state-of-the-art practices, making it more suited for validation through expert reviews and industry feedback rather than through controlled experiments or simulation scenarios. 

As such, while our approach offers a robust theoretical foundation for classifying and understanding autonomous vehicle systems, we want to demonstrate its suitability for the intended purpose. We therefore verify the applicability of the developed taxonomy to current AD systems by applying it to selected examples for current and future ADS.

\textbf{Example 1 - Truck highway pilot:} The first example is a theoretical but soon to be expected case of an automated truck capable of driving day and night between selected hubs linked by ordinary highways, without a driver. The SAE level therefore is level 4. The hubs might be located outside cities or towns and the services are limited to southern states in the US where rain but neither ice nor snow has to be expected. Using the terms of Table 1, the ODD would be described as follows: 4 $\vert$ US $\vert$ $\star$ $\vert$ H+ $\vert$ NR $\vert$ v4 $\vert$ ADRL6. The system is given an estimated ADRL of 6, as companies like Torc \cite{Torc2023} are piloting on highways. 

\textbf{Example 2 - Valet Parking:} The next example focuses on an automated valet parking system in a dedicated area of a parking garage without access for human beings. This can be summarized as:  4 $\vert$ $\star$ $\vert$ A $\vert$ S $\vert$ $\star$ $\vert$ v0. Again, we have a Level 4 system (humans not required at all inside the car) but are now in a dedicated area for automated vehicles only, with special “roads” at very low speeds. As we are “inside” a building there is absolutely no need for restricting environmental conditions such as weather or sight. To the author's knowledge, there are no commercial applications – most likely, due to the reason that there are not enough vehicles equipped with such a system on the market to be an interesting business case. All known approaches demonstrated are in mixed garages, described as 4 $\vert$ $\star$ $\vert$ $\star$   $\vert$ S $\vert$ $\star$ $\vert$ v0 and estimated at ADRL 8 or 9.

\textbf{Example 3 - Highway Pilot:} Another use case already mentioned in the introduction is a Highway Pilot as introduced to the markets by Mercedes-Benz, which has obviously an ADRL of 9. We would describe it as follows: 3 $\vert$ DE US $\vert$ $\star$ $\vert$ H $\vert$ LD $\vert$ v3 $\vert$ vehicleahead, noglare $\vert$ ADRL9. In our own test, we found, that the vehicle needs a vehicle driving in front of it, asks for hand-over prior to construction sites and as soon as it detects wet road conditions, cold temperature, glare, etc. 

\textbf{Example 4 - Robotaxis:} We describe the famous robotaxi services by Waymo in San Francisco as 4 $\vert$ US $\vert$ $\star$ $\vert$ H+U $\vert$ NR $\vert$ v3 $\vert$ onlySF $\vert$ ADRL9. They operate 24/7, meaning also at night and under rainy conditions. As the systems are in commercial use, they would be rated with ADRL 9. When applying the proposed approach and taxonomy, we discovered that any level 4 system for road type “H” does not make sense – as at least "H+" is required to get a sleeping passenger on a safe position at a parking place next to the highway.

\textbf{Example 5 - Mining trucks:} A very special case but one with several years of operation is the automated mining truck, which is on ADRL 9 and according to \cite{CAT} described briefly by: 4 $\vert$ $\star$ $\vert$ $\star$ $\vert$ S $\vert$ $\star$ $\vert$ v3 $\vert$ ADRL9.

\section{Discussion \& Conclusion}\label{chapter:discussion}
In this paper, we developed a new taxonomy to describe ADS using SAE Level and ODD description at an intermediate level of abstraction. In addition, a method to describe the current maturity of ADS was proposed using the ADRL as an adaptation of NASA's TRL. A number of current systems were described using the new taxonomy to demonstrate its applicability. In doing so, white spots became obvious, e.g. there is no commercial urban case in Germany at the moment.  

Given the diversity in autonomous vehicle designs, capabilities, and operational environments, we acknowledge that our model might generalize certain aspects. This could lead to a taxonomy that doesn't fully capture the nuances of all types of autonomous vehicles but, at the same time, reduces the complexity. Whilst the categories could be considered definitive, it is possible that the attributes will need to be sharpened. For example, future systems may be approved for specific precipitation value limits (e.g. light rain is included, strong rain is excluded), which is currently not mapped in the taxonomy. Therefore, future new systems and developments must be monitored and the ODD description format must be kept up to date. The use of theoretical methods may introduce subjectivity, especially in the interpretation of ODDs and TRLs. Different experts might have varying opinions on how a particular system should be classified. 

In the future, this new description format needs to be tested in practice. In particular, the extent to which the level of detail of the ODD description is useful for users must be checked. This has been deliberately kept at a medium level to facilitate usability. Therefore, the applicability of the ODD description to future trends in developments is to be checked.  Nevertheless, we now invite experts to use the proposed taxonomy for discussions and to provide feedback on its usefulness. With the proposed approach, we hope to have developed a language that will facilitate future technical discussions and comparisons of systems.

\bibliographystyle{IEEEtran}
\bibliography{literatur}
\end{document}